\documentclass[a4paper,fleqn]{caswenjia}
\usepackage[numbers]{natbib}
\usepackage{geometry}
\usepackage{amsmath}
\usepackage{amsfonts}
\usepackage{amssymb}
\usepackage{bbm}
\usepackage{hyperref}
\usepackage{caption}
\usepackage{mathtools} 

\usepackage{amsmath}

\def\tsc#1{\csdef{#1}{\textsc{\lowercase{#1}}\xspace}}
\tsc{WGM}
\tsc{QE}
\tsc{EP}
\tsc{PMS}
\tsc{BEC}
\tsc{DE}

\begin{document}
\begin{sloppypar}

\let\WriteBookmarks\relax
\def\floatpagepagefraction{1}
\def\textpagefraction{.001}
\shorttitle{DehazeMamba: SAR-guided Optical Remote Sensing Image Dehazing with Adaptive State Space Model}
\shortauthors{Zhicheng Zhao et~al.}

\title [mode = title]{DehazeMamba: SAR-guided Optical Remote Sensing Image Dehazing with Adaptive State Space Model}

\author[1,2]{Zhicheng Zhao}

\author[1]{Jinquan Yan}

\author[1,2]{Chenglong Li}
\cormark[1]
\ead{lcl1314@foxmail.com}

\author[1]{Xiao Wang}

\author[1]{Jin Tang}

\address[1]{Anhui Provincial Key Laboratory of Multimodal Cognitive Computation, Anhui University, Hefei 230601, China}
\address[2]{Information Materials and Intelligent Sensing Laboratory of Anhui Province, China}

\cortext[cor1]{Corresponding author}
\begin{abstract}
Optical remote sensing image dehazing presents significant challenges due to its extensive spatial scale and highly non-uniform haze distribution, which traditional single-image dehazing methods struggle to address effectively. While Synthetic Aperture Radar (SAR) imagery offers inherently haze-free reference information for large-scale scenes, existing SAR-guided dehazing approaches face two critical limitations: the integration of SAR information often diminishes the quality of haze-free regions, and the instability of feature quality further exacerbates cross-modal domain shift. To overcome these challenges, we introduce DehazeMamba, a novel SAR-guided dehazing network built on a progressive haze decoupling fusion strategy. Our approach incorporates two key innovations: a Haze Perception and Decoupling Module (HPDM) that dynamically identifies haze-affected regions through optical-SAR difference analysis, and a Progressive Fusion Module (PFM) that mitigates domain shift through a two-stage fusion process based on feature quality assessment. To facilitate research in this domain, we present MRSHaze, a large-scale benchmark dataset comprising 8,000 pairs of temporally synchronized, precisely geo-registered SAR-optical images with high resolution and diverse haze conditions. Extensive experiments demonstrate that DehazeMamba significantly outperforms state-of-the-art methods, achieving a 0.73 dB improvement in PSNR and substantial enhancements in downstream tasks such as semantic segmentation. The dataset is available at \url{https://github.com/mmic-lcl/Datasets-and-benchmark-code}.
\end{abstract}

\let\printorcid\relax
\maketitle

\section{Introduction}
Optical remote sensing image dehazing represents a critical low-level vision task aimed at restoring clear images from haze-degraded ones. This task presents unique challenges in the remote sensing domain due to the highly non-uniform distribution of haze across extensive spatial scales~\cite{10282275}. As optical remote sensing applications continue to expand into military, forestry, and agricultural domains, the degradation of image quality caused by haze significantly impairs the performance of various remote sensing interpretation algorithms. Given that approximately 50\% of the Earth's surface is covered by clouds at any given time, obtaining high-resolution, haze-free remote sensing imagery is essential for effective downstream analysis tasks~\cite{9191007}.

\begin{figure}[t]
  \centering
   \includegraphics[width=\linewidth]{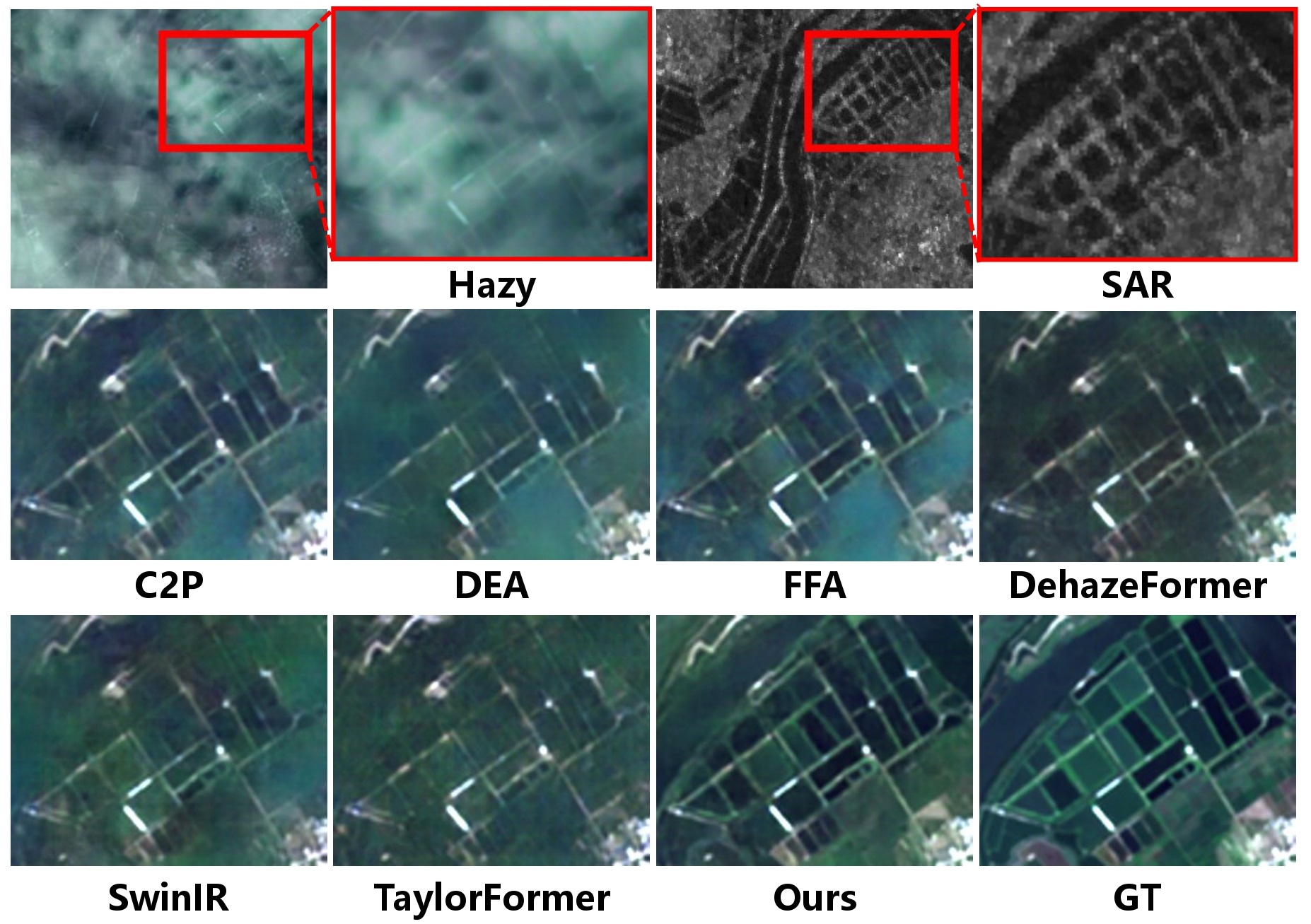}
   \caption{Comparative analysis of dehazing performance. DehazeMamba demonstrates superior texture recovery and detail preservation through SAR-guided dehazing, producing results that closely match the ground truth (GT) compared to state-of-the-art methods.}
   \label{data_example}
\end{figure}
As illustrated in Fig.~\ref{data_example}, remote sensing image dehazing remains a highly ill-posed problem due to the lack of haze-free reference images and the inherent uncertainty in the dehazing process~\cite{9892050}. Existing single-image dehazing methods, lacking reference information, often produce artifacts or fail to recover clear details when dealing with complex and variable remote sensing haze. Fortunately, Synthetic Aperture Radar (SAR) images are unaffected by haze and can provide valuable reference information for optical image dehazing, effectively addressing the issue of unstable images generated by single-image dehazing methods due to the lack of reference information~\cite{9211498,meraner2020cloud,ebel2022sen12ms}.

However, utilizing SAR images to guide the dehazing of optical remote sensing images presents two main challenges. First, due to the non-uniform characteristics of remote sensing haze~\cite{jiang2021deep}, there may be haze-free regions in the optical images where SAR information is not required. Introducing too much heterogeneous SAR information into these haze-free regions can disrupt the original pixel composition, causing image degradation and affecting the dehazing effect. Second, there is a significant domain shift between optical and SAR features, and direct fusion cannot sufficiently integrate the useful information in optical and SAR features. Moreover, due to hardware limitations and the influence of coherent imaging systems, targets in SAR images often contain noise~\cite{singh2016analysis}. Haze and noise cause the quality of optical and SAR features to be unstable, further increasing the difficulty of mitigating domain shift. Although some studies have attempted to improve dehazing results by integrating SAR information~\cite{9211498,meraner2020cloud,ebel2022sen12ms,grohnfeldt2018conditional,huang2020single}, they have not effectively addressed the aforementioned issues. These challenges hinder the performance of optical-SAR fusion dehazing algorithms, leading to suboptimal dehazed image quality. Therefore, solving these problems is crucial for improving image quality and advancing downstream tasks.

Due to the limitations of local receptive fields and quadratic computational complexity~\cite{xie2024fusionmamba,dong2024fusion,li2024mambadfuse}, current mainstream deep learning models, such as Convolutional Neural Networks CNNs and Transformers, exhibit noticeable deficiencies in the task of dehazing remote sensing images. First, since the haze in remote sensing images has unstable sizes, shapes, and densities, a larger receptive field is required to achieve effective dehazing results~\cite{10076399}. Second, as a pixel-level task that necessitates adjustment of each pixel, image dehazing is highly sensitive to computational costs, especially when processing high-resolution images~\cite{10378631}. Mamba~\cite{gu2023mamba}, as a novel architecture, can maintain a global receptive field while keeping linear computational complexity, making it very suitable for dehazing remote sensing images~\cite{fu2024hdmba,zheng2024u}.

In this work, we address the dual challenges of optical-SAR fusion for dehazing by proposing a novel progressive haze decoupling fusion strategy. This approach first decouples optical images into haze-affected and non-haze-affected regions, then mitigates domain shift in two progressive stages based on feature quality assessment, and adaptively fuses optical-SAR information specifically in haze-affected areas. Building on this strategy, we introduce DehazeMamba, a SAR-guided optical remote sensing image dehazing network utilizing the Mamba architecture. The network consists of two critical components: the Haze Perception and Decoupling Module (HPDM) and the Progressive Fusion Module (PFM). HPDM leverages the semantic differences between optical and SAR image pairs to dynamically identify and decouple haze-affected regions from optical imagery, while PFM implements a two-stage fusion process and effectively alleviates domain shift by using coarse-fusion features to guide fine-grained optical-SAR fusion-based on feature quality assessment.

To overcome the limitations of existing SAR-optical dehazing datasets, such as insufficient scale, inadequate resolution, and imprecise registration, we have constructed a comprehensive MRSHaze dataset. This resource includes 8,000 pairs of high-resolution, accurately registered SAR-optical images, encompassing a diverse range of haze concentrations across various scenes. Extensive experimental evaluation demonstrates that DehazeMamba significantly outperforms state-of-the-art methods, achieving a substantial improvement of 0.73 dB in PSNR and demonstrating notable enhancements in downstream applications such as semantic segmentation.

In summary, the main contributions of this paper are as follows:
\begin{itemize}

\item We propose DehazeMamba, a SAR-guided dehazing network with progressive haze decoupling fusion strategy for selective cross-modal feature integration.

\item To effectively identify and decouple haze-affected regions, we develop a Haze Perception and Decoupling Module (HPDM) that dynamically detects optical-SAR semantic differences and decouples features in haze-affected areas. Furthermore, to mitigate domain shifts between optical and SAR features, we propose a Progressive Fusion Module (PFM) that adaptively integrates cross-modal information based on feature quality in a progressive manner.

\item To facilitate research in SAR-guided optical remote sensing image dehazing, we introduce MRSHaze, a large-scale benchmark dataset containing 8,000 pairs of temporally synchronized, precisely geo-registered SAR-optical images with high resolution and diverse haze conditions.

\item Extensive experiments demonstrate DehazeMamba's superior performance in both dehazing quality and downstream tasks, validating the effectiveness of our proposed method.

\end{itemize}

\section{Related Work}
This section reviews the most relevant works, including methods for image dehaze, optical-SAR fusion, and state space models.

\subsection{Image Dehaze}

Early image dehazing approaches~\cite{5206515,berman2016non,ling2023single,wang2018novel,ju2021idrlp} were primarily based on the Atmospheric Scattering Model (ASM). The success of these methods lies in their incorporation of handcrafted priors into the ASM to estimate transmission and atmospheric light, thereby theoretically eliminating haze cover via the atmospheric model. For instance, the dark channel prior (DCP) proposed by He et al.~\cite{5206515} assumes that local patches in outdoor haze-free images contain pixels with minimal intensities in at least one color channel. Similarly, Ling et al.~\cite{ling2023single} developed a novel dehazing framework based on the saturation line prior. However, due to the limitations of these employed priors and underlying assumptions, such approaches often produce inaccurate estimations in complex scenarios, leading to unsatisfactory results.

Recent years have witnessed significant progress in deep learning-based dehazing methods, with various approaches leveraging CNNs and Transformers. Li et al.~\cite{8237773} and Chen et al.~\cite{8658661} proposed end-to-end CNN-based dehazing networks AOD-Net and GCA-Net, respectively, learning direct mappings between degraded and original images. Song et al.~\cite{10076399} introduced the first Transformer-based dehazing network, Dehazeformer, which leverages the global modeling capability of Transformers to achieve impressive dehazing effects on both natural and remote sensing images. With the recent emergence of Mamba, several studies have explored Mamba-based dehazing methods, achieving excellent results~\cite{fu2024hdmba,zheng2024u,zhou2024rsdehamba,ju2024improving}.

To capitalize on the advantages of multi-sensor data, researchers have proposed dehazing methods that leverage optical-infrared~\cite{YU2024128105} or optical-SAR fusion~\cite{9211498,meraner2020cloud,ebel2022sen12ms,grohnfeldt2018conditional,huang2020single}. Grohnfeldt et al.~\cite{grohnfeldt2018conditional} and Huang et al.~\cite{huang2020single} integrated optical and SAR images through conditional Generative Adversarial Networks (cGANs). Ebel et al.~\cite{9211498} employed a generative model to reduce the discrepancy between optical and SAR images, while Meraner et al.~\cite{meraner2020cloud} utilized a residual neural network to fuse optical-SAR information at the feature level.

While these methods have made significant contributions to image dehazing, challenges remain in addressing the highly non-uniform haze distribution in remote sensing imagery. DehazeMamba builds upon these advances by incorporating selective SAR information integration in haze-affected regions, while leveraging Mamba's efficient architecture to maintain global context awareness critical for effective remote sensing image dehazing.

\subsection{Optical-SAR Fusion}

Optical-SAR fusion methods have evolved into two primary categories: traditional approaches, such as transformation domain and spatial domain fusion algorithms, and deep learning-based methods. Traditional approaches rely on handcrafted feature extraction and mathematical transformations to integrate complementary information from multimodal data, whereas deep learning methods leverage hierarchical representations learned from large-scale datasets to achieve adaptive fusion.

Early fusion strategies predominantly utilized multi-resolution analysis (MRA) and component substitution (CS). For instance, wavelet transform-based methods~\cite{xiang2024two} decomposed optical and SAR images into frequency domains, selectively merging coefficients to enhance spatial details while preserving spectral fidelity. The intensity-hue-saturation (IHS) framework, adapted for SAR-optical fusion, replaced the intensity component with SAR data to emphasize structural features~\cite{2020IHS}. Additionally, principal component analysis (PCA) was employed to project multimodal features into orthogonal subspaces, though it often introduced spectral artifacts in heterogeneous regions~\cite{xiong2022robust}. While these methods achieved moderate success, they struggled with nonlinear relationships between modalities and required manual parameter tuning for specific scenarios. 

Deep learning-based methods can be divided into convolutional neural networks and generative adversarial networks. In CNN-based methods, Li et al.~\cite{9019866} pioneered the use of fusion networks with bilinear pooling layers to merge features extracted from optical and SAR images, exploring inter-channel relationships through attention mechanisms. Feng et al.~\cite{feng2019integrating} proposed branch CNN architectures to separately extract features from SAR and optical images. In GAN-based approaches, Gao et al.~\cite{8866939} transformed SAR images into optical representations for cloud area completion, providing a novel approach for heterogeneous image fusion. Fu et al.~\cite{fu2021reciprocal} introduced a multi-level cascaded residual connection GAN framework to achieve bidirectional transformation between optical and SAR images.

These fusion strategies have established valuable foundations for multimodal remote sensing applications. DehazeMamba extends this progress by introducing a progressive fusion approach that considers the varying quality of optical and SAR features across different image regions. This adaptive integration helps preserve detail integrity in non-hazy areas while enhancing information recovery in haze-obscured regions.

\subsection{State Space Model}
 State Space Models (SSMs) have recently demonstrated remarkable effectiveness in capturing the dynamics and dependencies of sequential data through state space transformations. Gu et al.~\cite{gu2021efficiently} proposed a novel structured state-space sequence model (S4) that applies low-rank structural correction to effectively handle long-range dependencies in sequence data. Smith et al.~\cite{smith2022simplified} introduced S5, which incorporates multi-input multi-output (MIMO) SSMs and parallel scan algorithms. Gu et al.~\cite{gu2023mamba} further developed S6, employing data-dependent SSM layers that balance efficiency and effectiveness while maintaining linear computational complexity. Inspired by these advancements, Zhu et al.~\cite{zhu2024vision} achieved linear complexity without sacrificing global receptive field capabilities, effectively addressing the quadratic complexity limitations of Vision Transformers (ViTs).


These advancements in State Space Models have demonstrated promising capabilities for sequence modeling tasks. Our work incorporates these architectural innovations to the specific challenges of remote sensing image dehazing, where DehazeMamba leverages the strengths of SSMs to effectively capture spatial dependencies across modalities while maintaining computational efficiency when processing high-resolution imagery.

\section{Method}
DehazeMamba enhances optical remote sensing image dehazing by leveraging complementary SAR imagery. The network employs two key innovative modules: the Haze Perception and Decoupling Module (HPDM) for identifying and isolating haze-affected regions, and the Progressive Fusion Module (PFM) for effective integration of RGB-SAR features to guide the dehazing process.

\begin{figure*}    
    \centering
    \includegraphics[width=1\textwidth]{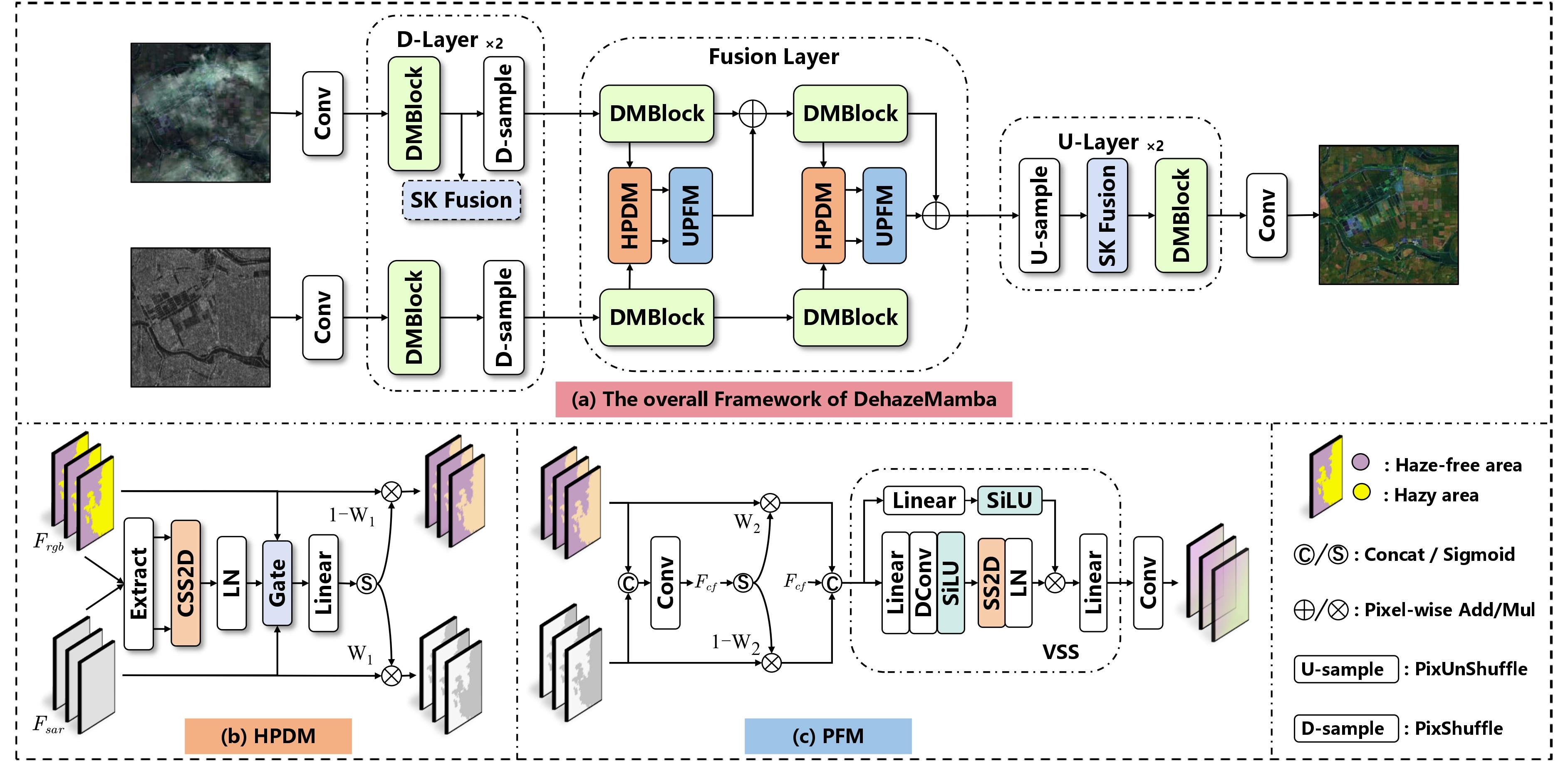}
    \captionsetup{labelfont={color=blue}}
    \caption{
    Architecture overview of DehazeMamba. The network comprises dual-branch encoders that process hazy optical and SAR images separately to extract multi-level features. These features undergo fusion through the HPDM and PFM modules in the Fusion Layer. The decoder then reconstructs the fused features into a haze-free output image, with residual connections preserving spatial details across different scales.
    }
    \label{framework}
\end{figure*}

\subsection{Overall Framework}
\label{A}
As illustrated in Fig.~\ref{framework}(a), DehazeMamba employs a dual-branch Mamba-based encoder to extract hierarchical features from both optical and SAR inputs. Each branch initially adjusts channel dimensions through convolutional layers and implements progressive down-sampling to capture multi-scale information. The high-level optical ($R_{i}$) and SAR ($S_{i}$) features undergo processing through specialized DMBlocks, followed by integration in the HPDM and PFM modules to produce fused representations $F_{i}$. These fused features are subsequently combined with $R_{i+1}$ for further refinement in the optical branch. The decoder architecture employs up-sampling operations to progressively recover spatial information while reducing channel dimensions, culminating in a final convolutional layer that generates the dehazed image. To mitigate information loss during processing and enhance overall stability, we incorporate residual connections between corresponding down-sampling and up-sampling layers, employing SK Fusion~\cite{8954149} to effectively merge features across different levels of the network. 

In the stage of fusing high-level optical and SAR features, in order to specifically fuse optical-SAR information in the haze areas and reduce the impact on non-haze areas, we first use the HPDM to accurately identify the areas affected by haze in the optical features. Subsequently, with the aim of alleviating the problem of substandard fusion outcomes resulting from domain shift and attaining enhanced dehazing effects, the PFM is utilized to perform the two-stage fusion of optical-SAR features in strict accordance with the feature quality.

\subsection{Haze Perception and Decoupling Module}
\label{B}
The HPDM identifies haze-affected regions by detecting informational discrepancies between optical and SAR feature representations. Accurate perception of these differences is essential for precise haze region decoupling. While existing approaches utilizing CNNs and Transformers demonstrate effectiveness in recognizing simple image disparities, they often struggle with widely distributed and complex spatial differences that characterize haze patterns in remote sensing imagery. The limited receptive field of CNNs constrains their ability to capture extensive and intricate differences between optical and SAR features. Transformers, meanwhile, emphasize sequence element relationships over spatial positions, resulting in reduced spatial awareness that hinders their performance in extracting spatially nuanced differences. In contrast, Mamba employs a global receptive field and models multiple sequence directions, significantly enhancing spatial awareness and enabling effective extraction of spatially distributed and complex morphological differences between optical and SAR features.

As depicted in Fig.~\ref{framework}(b), HPDM is a Mamba-based module comprising two primary processes: haze perception and image decoupling. During haze perception, HPDM extracts modality-specific difference information between optical and SAR features and transforms this information into a weight map $W1$ through a Sigmoid function, representing the spatial distribution and concentration of haze. In the decoupling phase, to isolate SAR information corresponding to haze-affected regions and attenuate haze information in optical features, HPDM applies $W1$ to the SAR features and $(1-W1)$ to the optical features, respectively, generating the final output through pixel-wise multiplication operations.

In HPDM, the two input feature maps initially undergo individual processing through an 'Extract' operation. This process can be formulated as:
\begin{equation}
\begin{split}
x _{i}^{out}=SiLU(DConv(Linear(x _{i}^{in})), 
\end{split}
\end{equation}
where index $i$ represents either 'rgb' or 'sar'. The processed modal features are then input into the CSS2D module for modality-specific feature extraction. Within CSS2D, linear layers process the input features to generate essential parameter matrices: $\Delta_{rgb}$, $A_{rgb}$, $B_{rgb}$, $C_{rgb}$, $D_{rgb}$ for RGB, and $\Delta_{sar}$, $A_{sar}$, $B_{sar}$, $C_{sar}$, $D_{sar}$ for SAR. Matrices $A$ and $B$ require discretization prior to their application in state space equations, formulated as:
\begin{align}
\MoveEqLeft \bar{A} _{rgb}=\exp  (\Delta _{rgb}A_{rgb}), \\
\MoveEqLeft \bar{A} _{sar}=\exp  (\Delta _{sar}A_{sar}), \\
\MoveEqLeft \bar{B} _{rgb}=\Delta _{rgb}B_{rgb}, \\
\MoveEqLeft \bar{B} _{sar}=\Delta _{sar}B_{sar},
\end{align}
here, $\Bar{M}_{x}$ denotes the discretized matrix $M$ of modality $x$. The discretized matrices $\Bar{A}_{rgb}$, $\Bar{B}_{rgb}$, $\Bar{A}_{sar}$, and $\Bar{B}_{sar}$ are utilized to iterate over the hidden state $h_{k}$ of their respective modalities according to:
\begin{align}
h_{rgb}^{t} = \bar{A} _{rgb}h_{rgb}^{t-1}+\bar{B}_{rgb}x_{rgb}^{t}, \\
h_{sar}^{t} = \bar{A} _{sar}h_{sar}^{t-1}+\bar{B}_{sar}x_{sar}^{t}, 
\end{align}
where $x_{rgb}^{t}$ and $x_{sar}^{t}$ represent the inputs of optical and SAR modes at time step $t$, corresponding to the CSS2D input features. 

Drawing inspiration from Han et al.~\cite{Han_2018_ECCV}, we derive a difference-perceiving $C$ matrix by computing the absolute difference between $C_{rgb}$ and $C_{sar}$. This matrix decodes features representing modality-specific information from both modalities' hidden states:
\begin{align}
\MoveEqLeft {\color{red} C} = \left | C _{rgb}-C _{sar} \right |,\\
\MoveEqLeft y_{rgb}^{t} = {\color{red} C}h_{rgb}^{t}+D_{rgb}x_{rgb}^{t},\\
\MoveEqLeft y_{sar}^{t} = {\color{red} C}h_{sar}^{t}+D_{sar}x_{sar}^{t}
\end{align}

To further accentuate the differing features, the absolute difference between the decoded features from each modality is taken as the CSS2D output:
\begin{equation}
\begin{split}
y_{out}^{t}  = \left | y_{rgb}^{t}-y_{sar}^{t} \right |
\end{split}
\end{equation}

To enhance information flow control, we implement a joint gate mechanism ('Gate' operation) wherein optical and SAR features collaboratively provide gating information. The two HPDM inputs undergo separate linear transformations before being summed and processed through a SiLU activation function to generate joint gating information that adaptively modulates feature flow.

\subsection{Progressive Fusion Module}
\label{C}
When integrating optical and SAR features, a significant domain shift challenge arises, as direct fusion often fails to effectively leverage useful information from both modalities. Moreover, haze in optical features and noise in SAR features further destabilize feature quality, exacerbating domain shift issues. Existing methods typically address either modality quality or domain shift in isolation, without considering their interrelated nature. To overcome these limitations, we propose the Progressive Fusion Module (PFM), which addresses domain shift through modality quality adaptation in two sequential stages, enabling more effective integration of optical and SAR information.

In optical-SAR feature integration, both SAR feature noise and optical feature haze affect dehazing difficulty. For instance, even with minimal SAR noise, introducing SAR information may increase dehazing complexity if the optical features exhibit only light haze. Conversely, when haze severely obscures optical features, incorporating SAR information, despite higher noise levels, can significantly reduce dehazing difficulty. PFM progressively fuses optical-SAR features in two stages based on information quality assessment to maximize dehazing effectiveness, as illustrated in Fig.~\ref{framework}(c).

Let $F_{o}$ and $F_{s}$ represent the optical and SAR features adjusted by PFM, respectively. In the first stage, $F_{o}$ and $F_{s}$ are concatenated along the channel dimension, and channel dimensionality is adjusted through convolutional operations to generate a coarse fused feature $F_{cf}$. This coarse fusion feature serves dual purposes: comprehensively assessing optical and SAR feature quality while providing guidance for subsequent, more refined optical-SAR fusion. In the second stage, to adaptively adjust feature weights according to modality quality, a Sigmoid function transforms $F_{cf}$ into a weight map $W_{2}$. After allocating $W_{2}$ and $(1-W_{2})$ to $F_{o}$ and $F_{s}$ respectively, pixel-wise multiplication generates adjusted features $F_{ao}$ and $F_{as}$. 

To alleviate domain discrepancies, $F_{cf}$ provides essential guidance information for optical-SAR feature fusion. After concatenating $F_{ao}$, $F_{as}$, and $F_{cf}$ along the channel dimension, we employ a Vision State Space (VSS)~\cite{zhu2024vision} module followed by convolutional operations for fine-grained fusion, yielding the final output feature representation $F_{out}$.

\begin{figure}[!t]
\centering
\includegraphics[width=0.8\linewidth]{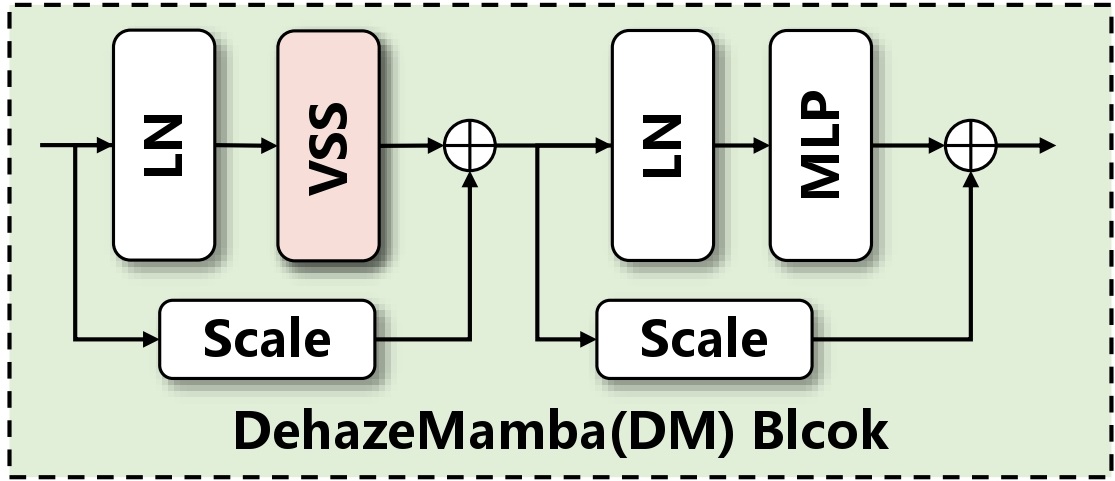}
\caption{Architectural details of the DehazeMamba (DM) Block. The module combines VSS for global context modeling with MLP for local feature enhancement, incorporating residual connections with learnable parameters to facilitate efficient information flow.}
\label{dmblock}
\end{figure}

\subsection{DehazeMamba(DM) Block}
The DMBlock serves as the core feature extraction component of the backbone network, designed to effectively capture both global contextual information and local details. This is achieved through a strategic combination of Vision State Space (VSS)~\cite{zhu2024vision} and Multi-Layer Perceptron (MLP) components, as illustrated in Fig.~\ref{dmblock}. VSS effectively extracts global contextual information, while MLP enhances local feature details. To ensure efficient information propagation throughout the network, DMBlock incorporates residual connections with learnable parameter controls between the input and output of both VSS and MLP stages, allowing adaptive adjustment of information flow based on feature characteristics.

\subsection{Loss Function}
To constrain the image generation process comprehensively across different domains, we employ L1 loss functions in both spatial and frequency domains:
\begin{align}
\MoveEqLeft \mathcal{L}_{spatial} = \lVert \hat{Y}_{s} - Y_{s} \rVert_{1}, \\
\MoveEqLeft \mathcal{L}_{frequency} = \lVert \mathcal F(\hat{Y}_{s})-\mathcal F(Y_{s}) \rVert_{1},  
\end{align}
where $\mathcal F$ denotes the fast Fourier transform, $\hat{Y}_{s}$ represents the output image, and $Y_{s}$ the target image. The final loss function is defined as $\mathcal{L}=\mathcal{L}_{spatial}+\lambda \mathcal{L}_{frequency}$, with $\lambda$ set to 0.1.

\section{Dataset}
This section introduces MRSHaze, a comprehensive benchmark dataset specifically designed for SAR-guided optical remote sensing image dehazing. We present the motivation behind its creation and detail the data collection process.


\begin{table}[!t]
\renewcommand{\arraystretch}{1.2}
\begin{center}
\caption{Detailed comparison between different datasets. W-A represents whether the optical or optical-SAR image pairs in the dataset are well-aligned.  DN represents whether the SAR image has been denoised }
\begin{tabular}{ p{2.4cm} >{\centering\arraybackslash}p{0.5cm} >{\centering\arraybackslash}p{0.7cm} >{\centering\arraybackslash}p{1.6cm}   >{\centering\arraybackslash}p{0.6cm} >{\centering\arraybackslash}p{0.5cm}}
\hline
Dataset& SAR  &W-A& Resolution   & Num     & DN \\
\hline
O-HAZE~\cite{ancuti2018haze} & $ \times $ & $\surd$ & 1280x720   & 45 & -\\
I-HAZE~\cite{ancuti2018ihaze} & $ \times $ & $\surd$ & 4657×2833 &  35  & -\\
Dense-Haze~\cite{ancuti2019dense} & $ \times $ & $\surd$ & 1600×1200 &  33 & -\\
Haze4k~\cite{liu2021synthetic} & $ \times $ & $\surd$ & 400×400  &  4K & -\\
RS-Haze~\cite{10076399} & $ \times $ & $\surd$ & 512×512 &  52K & -\\
RESIDE~\cite{li2019benchmarking} & $ \times $ & $\surd$ & 460×620  & 45K & -\\
SEN12MS-CR~\cite{9211498} & $\surd$  & $ \times $ & 256×256   & 120K & $ \times $\\
SateHaze1k~\cite{huang2020single}& $\surd$  & $ \times $ & 512×512  &  1K & $ \times $\\
MRSHaze(Ours) & $\surd$ & $\surd$&512×512   &  8K   &  $\surd$\\
\hline
\label{tab:dataset}
\end{tabular}
\end{center}
\vspace{-1.0cm}
\end{table}

\begin{figure}[!b]
  \centering
   \includegraphics[width=\linewidth]{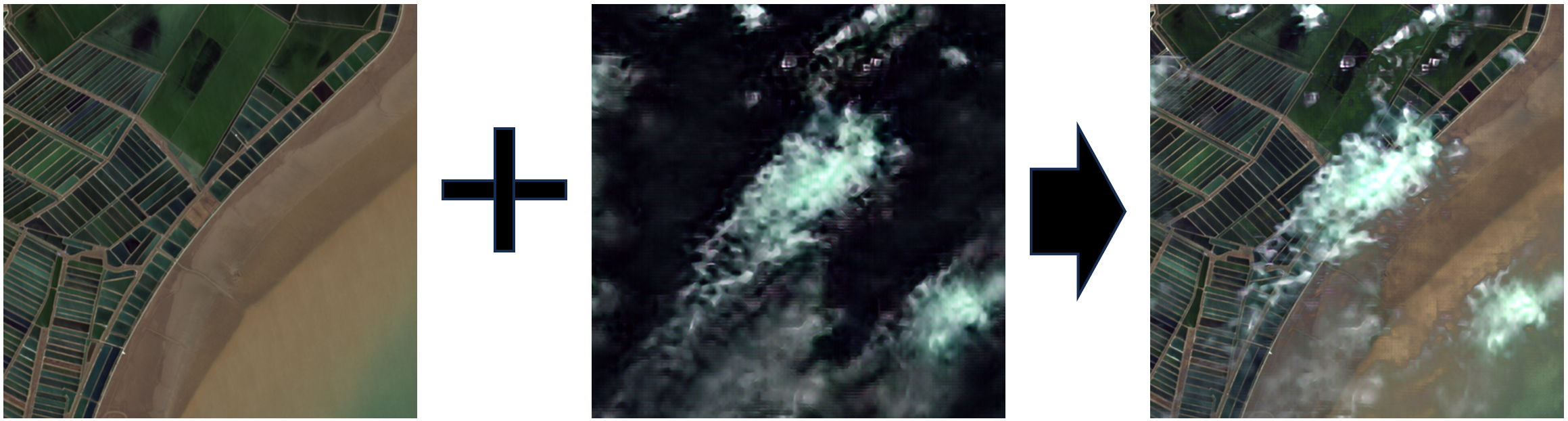}
   \caption{Synthetic haze generation process. We combine clear optical remote sensing images from Sentinel-2 L2A Product with realistic haze patterns derived from cloud masks extracted from Sentinel-3 SLSTR Product and transformed via a pre-trained CycleGAN model.}
   \label{add}
\end{figure}

\subsection{Motivation for MRSHaze}
The development of effective remote sensing image dehazing algorithms has been significantly hindered by limitations in existing datasets, particularly in addressing the challenges of widespread and non-uniform haze distribution. The SEN12MS-CR dataset~\cite{9211498} suffers from insufficient resolution (256×256 pixels), excessive noise in SAR imagery, and temporal misalignment in optical image acquisition. Similarly, the SateHaze1k dataset~\cite{huang2020single} is constrained by limited data scale (1,200 pairs) and inadequate alignment between optical and SAR data. To overcome these limitations, we introduce MRSHaze, a high-resolution (512×512 pixels) dataset that offers:
\begin{itemize}
\item Enhanced representation of large-scale haze distribution, facilitating the development of models with improved receptive fields. 
\item Precise SAR-optical alignment using geographical information, enabling more effective fusion techniques. 
\item A comprehensive benchmark for evaluating SAR-guided optical image dehazing algorithms in complex real-world scenarios. 
\end{itemize}
MRSHaze thus addresses critical gaps in existing resources, providing a robust foundation for advancing the field of multimodal remote sensing image dehazing.

\subsection{Data Collection and Analysis}
The MRSHaze dataset comprises 8,000 pairs of high-resolution (512×512 pixels) SAR and optical images derived from Sentinel-1 and Sentinel-2 satellites, respectively. SAR data, acquired in Interferometric Wide Swath mode with VH polarization, undergoes rigorous preprocessing including orthorectification, multi-view processing, and noise reduction using the BM3D algorithm~\cite{lin2021improved}. Optical data utilizes RGB channels (bands 2, 3, 4) and is carefully selected from tiles with cloud coverage not exceeding 0.5\%. Both modalities are precisely aligned to ensure optimal spatial correspondence.

Hazy images are synthetically generated by combining clear optical images with realistic cloud patterns. The dataset captures diverse scenes from global locations, including urban, rural, and natural landscapes under various environmental conditions. For benchmarking purposes, the dataset is divided into 6,000 training pairs and 2,000 testing pairs, maintaining a 3:1 ratio to ensure robust model evaluation.

This comprehensive dataset addresses the limitations of existing resources, providing a robust foundation for developing and evaluating SAR-guided optical image dehazing algorithms in complex, real-world scenarios.

\begin{figure}[!b]
  \centering
  
   \includegraphics[width=\linewidth]{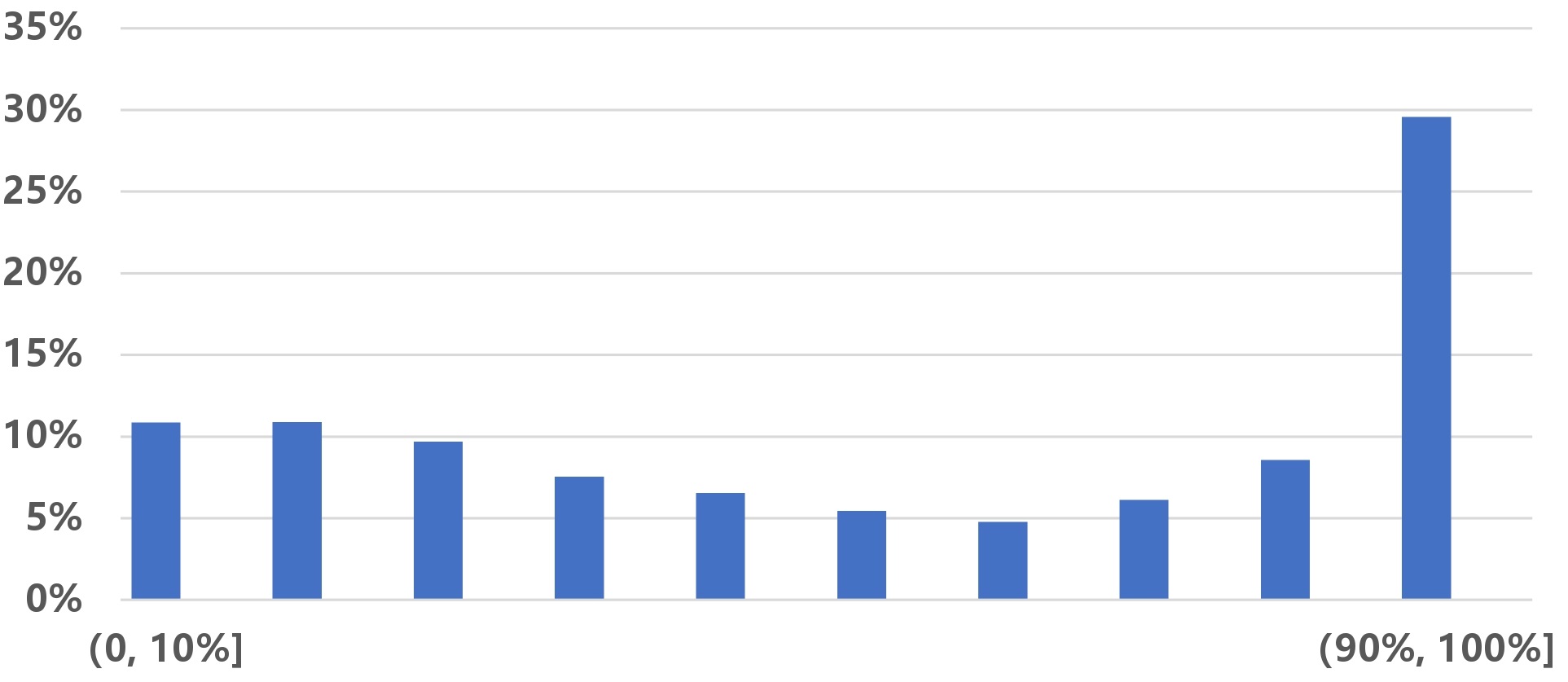}
   \caption{Statistical analysis of haze region proportions in the MRSHaze dataset. The horizontal axis represents percentage intervals of haze coverage per image, while the vertical axis shows the proportion of dataset images falling within each interval. The dataset exhibits a balanced distribution across different haze coverage ranges.}
   \label{haze_distribution}

   \vspace{0.3cm}

   \includegraphics[width=\linewidth]{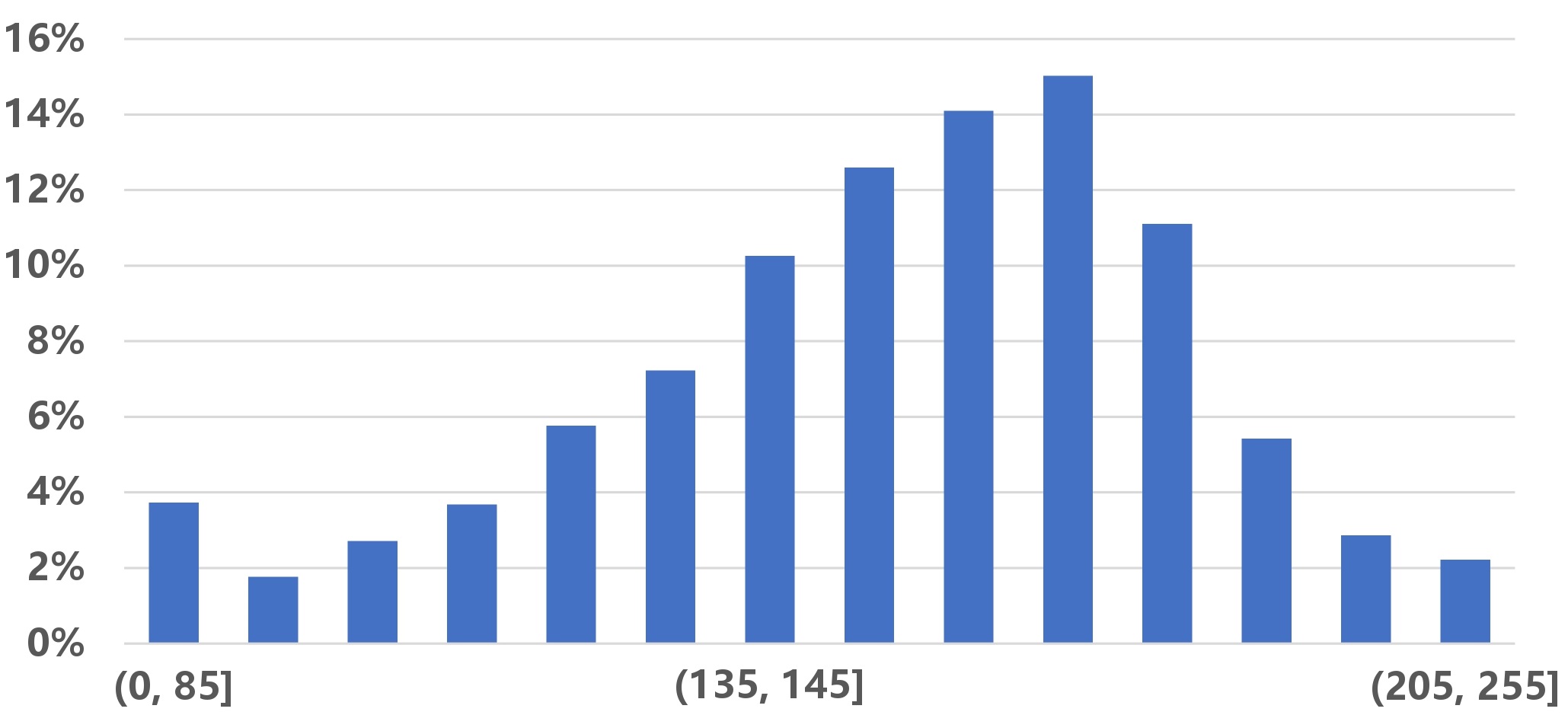}
   \caption{Analysis of haze density distribution in the MRSHaze dataset. The horizontal axis represents density intervals of haze intensity, while the vertical axis indicates the proportion of dataset images within each density range. The dataset contains a comprehensive spectrum of haze densities from light to heavy.}
   \label{haze_density}
\end{figure}

\subsection{Adding Haze}
As shown in Fig.~\ref{add}, we generate synthetic hazy images by overlaying remote-sensing haze patterns onto clear optical images. The clear images are sourced from the Sentinel-2 L2A Product. For the haze component, we first extract cloud masks from the Sentinel-3 SLSTR Product, then transform these masks into realistic haze patterns using a pre-trained CycleGAN~\cite{CycleGAN2017} model, ensuring the synthetic haze accurately mimics the characteristics of natural atmospheric haze in remote sensing imagery.

\begin{figure}[!b]
  \centering

   \includegraphics[width=0.7\linewidth]{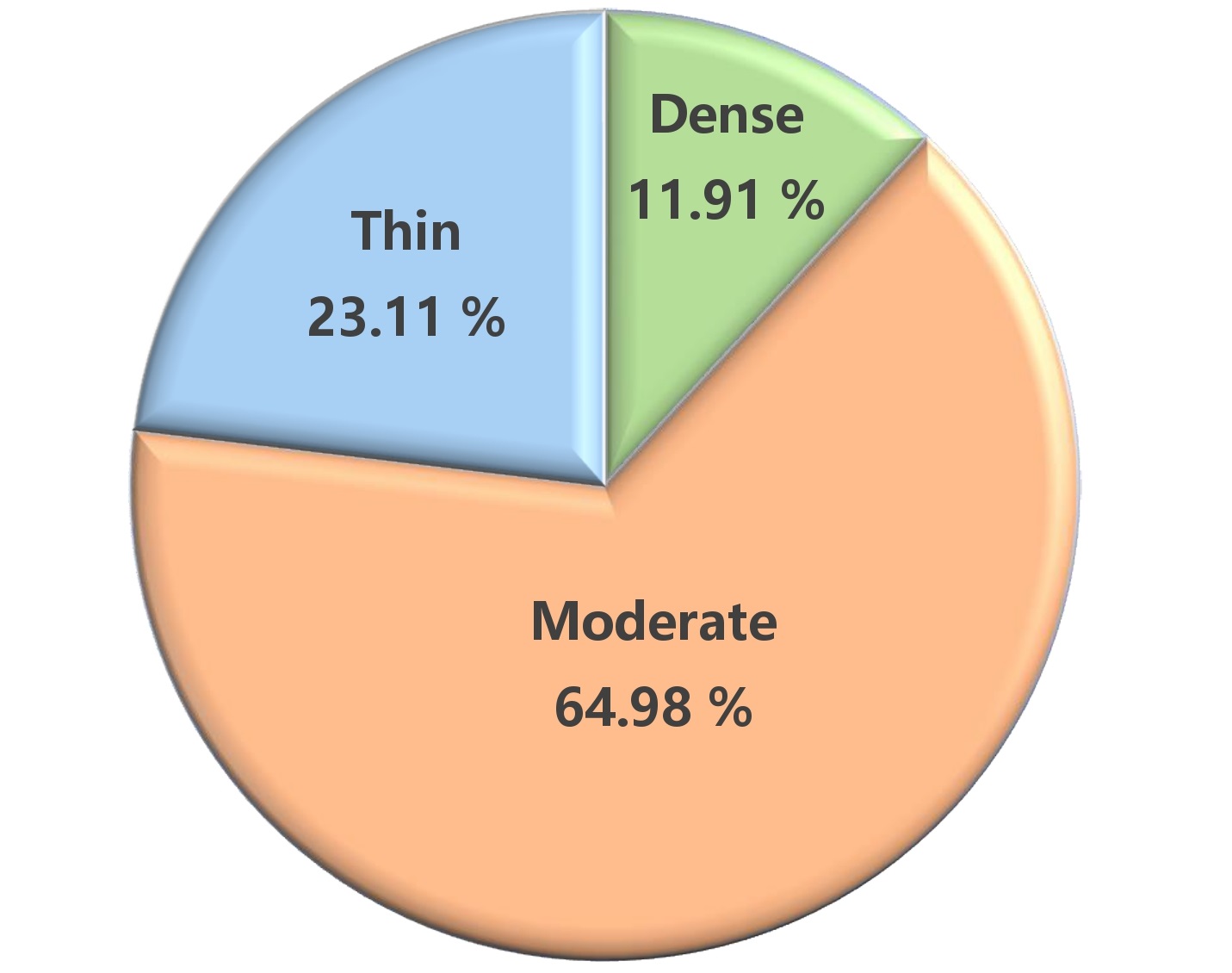}
   \caption{Proportional distribution of three haze concentration levels in MRSHaze: thin (23.11\%), moderate (64.98\%), and dense (11.91\%). This balanced distribution ensures comprehensive algorithm evaluation across varying haze severity.}
   \label{density_ratio}

   \vspace{0.3cm}
   
   \includegraphics[width=\linewidth]{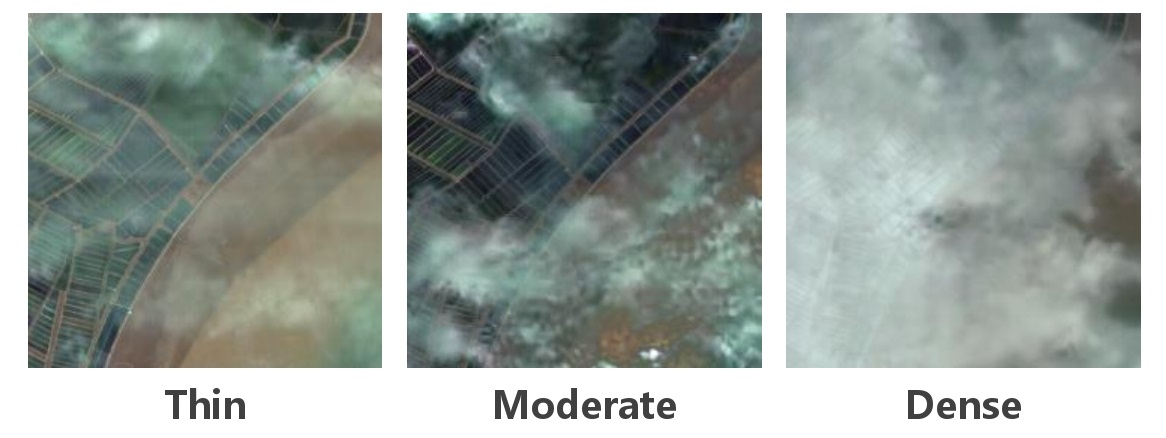}
   \caption{Visual examples of haze images at three concentration levels: thin (left), moderate (center), and dense (right), illustrating the range of haze conditions represented in the MRSHaze dataset.}
   \label{density_visualization}
   
\end{figure}

\subsection{Distribution and Density of Haze}

\label{sec:Haze}
To enhance the diversity and representativeness of haze patterns in MRSHaze, we incorporate RS haze with varying distributions and densities when generating synthetic hazy images.

We conducted a comprehensive statistical analysis of haze region proportions across the MRSHaze dataset, where the proportion refers to the percentage of haze-affected area within each image. As illustrated in Fig.~\ref{haze_distribution}, the dataset exhibits a broad distribution of haze coverage, with each 10\% interval from 0\% to 100\% containing between 4.77\% and 29.58\% of the total images. This diverse distribution necessitates dehazing algorithms capable of effectively handling haze regions of varying sizes and spatial arrangements.

To quantify haze concentration, we utilize the average value of the brightest 30\% of pixels within haze-affected regions. As shown in Fig.~\ref{haze_density}, the dataset demonstrates a comprehensive distribution across various haze concentration intervals, particularly spanning from light to heavy haze conditions. This diverse concentration range imposes stringent requirements on the robustness and adaptability of dehazing algorithms, facilitating more thorough evaluation of their performance across different atmospheric conditions.

Based on the average value of the brightest 30\% of pixels in haze regions, we categorize three distinct haze concentration levels: thin (value $\leq$ 105), moderate (105  $<$ value $\leq$ 175), and dense (value $>$175). Fig.~\ref{density_ratio} illustrates the distribution proportions of these three categories, while Fig.~\ref{density_visualization} provides visual examples of each concentration level, demonstrating the dataset's comprehensive coverage of haze conditions.

\begin{figure*} 
\centering
\includegraphics[width=\textwidth]{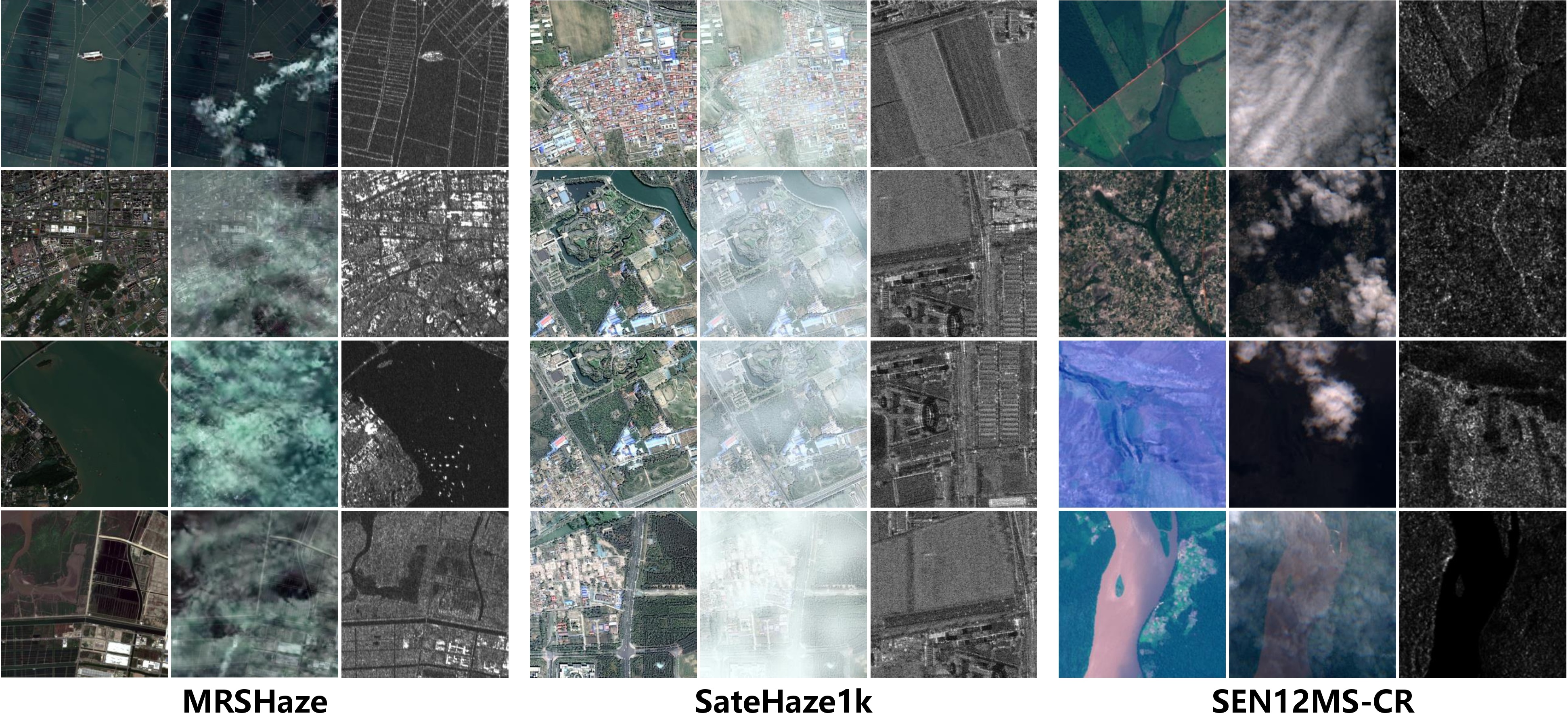}
\caption{Visual comparison of sample pairs from MRSHaze, SateHaze1k, and SEN12MS-CR datasets. MRSHaze exhibits superior alignment between optical and SAR images with notably higher SAR image quality. SateHaze1k demonstrates evident misalignment between modalities, while SEN12MS-CR shows color distortion in optical images and contains more noise in SAR imagery.}
\label{data_compare}
\end{figure*}

\subsection{Comparison with Other Datasets}
Table~\ref{tab:dataset} presents a comparative analysis of MRSHaze against established dehazing datasets. Prior to MRSHaze, two primary optical-SAR fusion dehazing benchmark datasets existed: SEN12MS-CR and SateHaze1k. SEN12MS-CR is limited by low image resolution (256×256) and poor temporal alignment of optical imagery. SateHaze1k suffers from misalignment between optical and SAR components and restricted data volume (1K pairs). In contrast, MRSHaze provides 8,000 precisely aligned, high-resolution (512×512) image pairs, effectively addressing the limitations of both predecessors. Furthermore, the SAR imagery in MRSHaze undergoes specialized denoising treatment—a critical preprocessing step absent in the other two datasets. Representative samples from MRSHaze, SEN12MS-CR, and SateHaze1k are displayed in Fig.~\ref{data_compare}, visually demonstrating the qualitative improvements in our dataset.

\section{Experiments}

\begin{figure*}  
\begin{center}

\includegraphics[width=\textwidth]{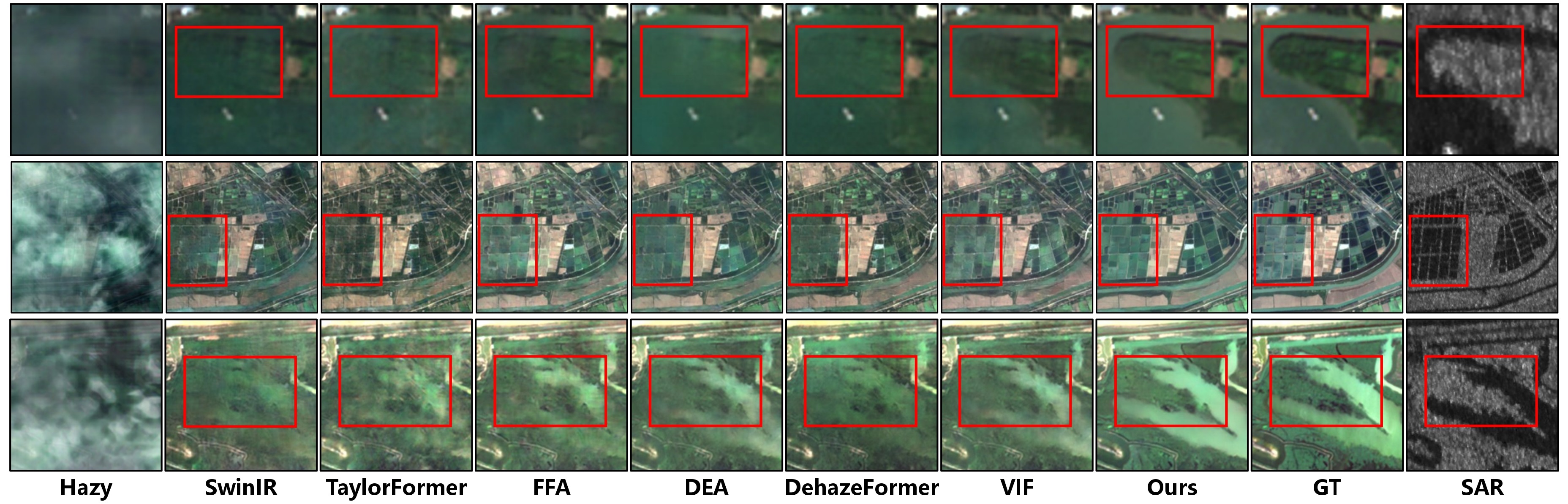}
\caption{Qualitative comparison on the MRSHaze dataset. DehazeMamba achieves superior texture recovery and detail preservation with fewer artifacts compared to competing methods. Single-image methods and existing guided approaches demonstrate significant limitations in texture reconstruction and effective SAR information utilization, particularly in heavily haze-affected regions.}
\label{visualization}

\vspace{0.3cm}

\includegraphics[width=\textwidth]{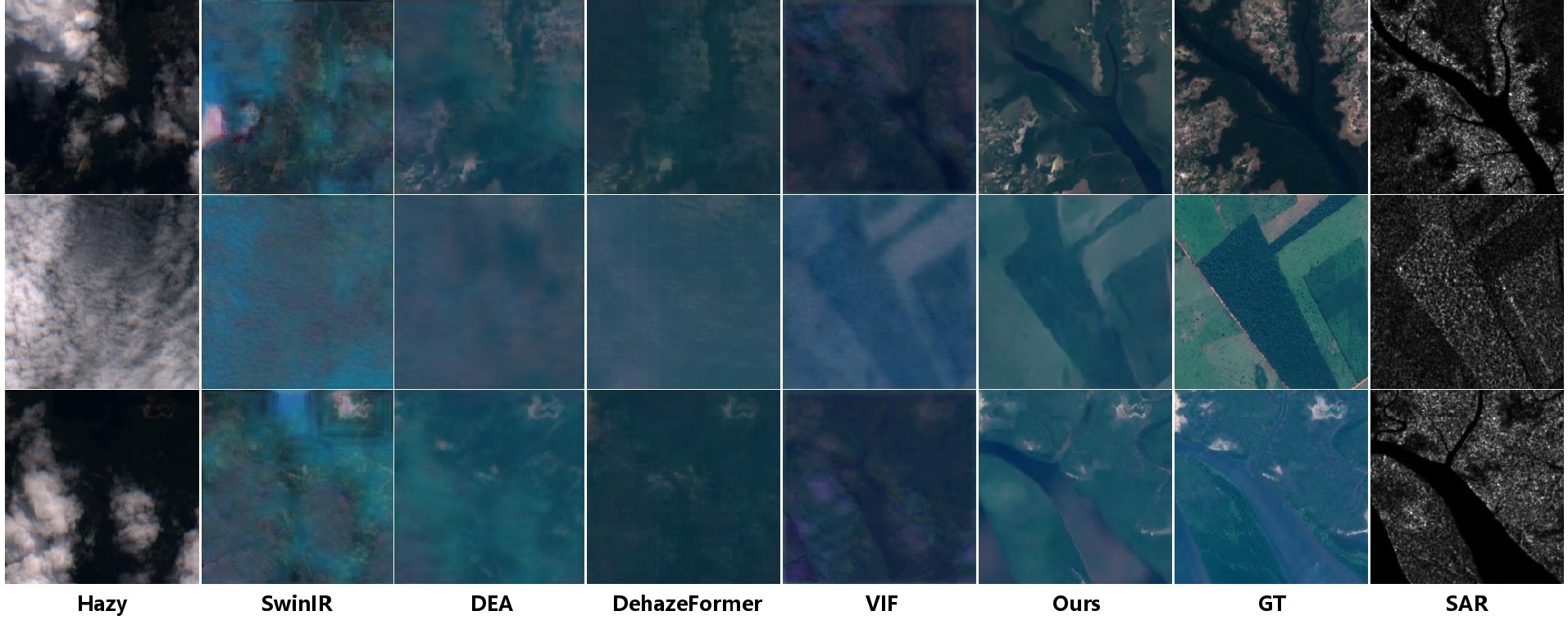}
\caption{Qualitative comparison on the SEN12MS-CR dataset. DehazeMamba demonstrates superior color fidelity and texture reconstruction compared to both single-image and existing guided dehazing methods, which exhibit notable distortion and information loss across various scene types.}
\label{cr1}

\vspace{0.3cm}

\includegraphics[width=\textwidth]{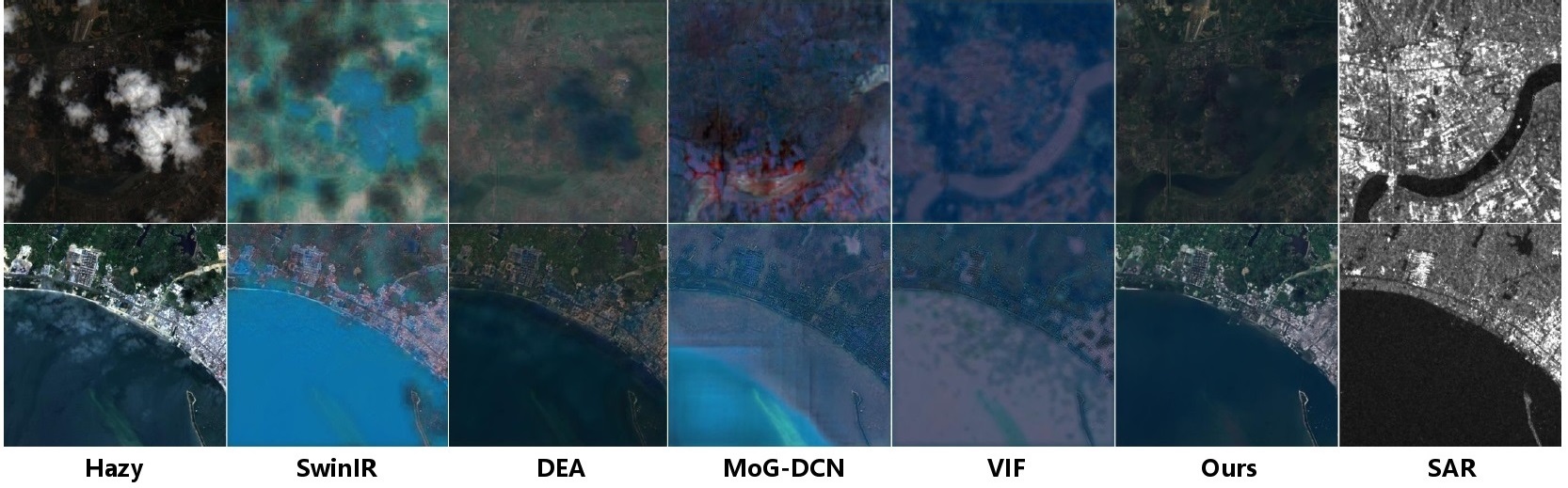}
\caption{Qualitative comparison on real-world hazy images. DehazeMamba successfully recovers information in haze-obscured regions while preserving detail integrity in haze-free areas, producing results with more natural color reproduction compared to alternative approaches.}
\label{real}

\end{center}
\end{figure*}

We conduct extensive experiments to evaluate the efficacy of DehazeMamba. This section presents our experimental methodology, quantitative and qualitative analyses, and ablation studies. 

\subsection{Experimental Settings}
\noindent\textit{1) Implementation Details}: DehazeMamba is implemented in PyTorch and trained on an NVIDIA RTX 4090 GPU. We define three model scales: DehazeMamba-T, -B, and -L, with [2, 2, 2, 1, 1], [4, 4, 4, 2, 2], and [8, 8, 8, 4, 4] blocks at respective network stages [N1, N2, N3, N4, N5]. We employ the AdamW optimizer with an initial learning rate of 2e-4 and batch size of 6. The learning rate decreases from 2e-4 to 1e-6 following a cosine annealing schedule. Our loss function combines L1 losses in both spatial and frequency domains, with the frequency domain loss weighted at 0.1.

\noindent \textit{2) Comparison Methods and Metrics}: We evaluate DehazeMamba against state-of-the-art methods on both MRSHaze and SEN12MS-CR datasets. The comparison includes:
\begin{itemize}
\item[a)]Single-modal methods: SwinIR~\cite{9607618}, FFA-Net~\cite{qin2020ffa}, DehazeFormer~\cite{10076399}, MixDehazeNet~\cite{Mixdehazenet}, C2PNet~\cite{Zheng_2023_CVPR}, DEA~\cite{10411857}, SF-Net~\cite{cui2023selective}, FS-Net~\cite{10310164}, TaylorFormer~\cite{10378631} and FocalNet~\cite{Cui_2023_ICCV}. 
\item[b)]Multi-modal methods:VIF~\cite{YU2024128105}, UGSR~\cite{gupta2021toward}, and MoG-DCN~\cite{dong2021model}. Additionally, we adapt DEA and FocalNet to multi-modal approaches (GDEA and GFocalNet) by integrating a dual-branch encoder and cross-attention fusion
\end{itemize}
Performance is assessed using Peak Signal-to-Noise Ratio (PSNR) and Structural Similarity Index (SSIM) metrics. For single-modal dehazing methods, we train using paired optical images from the datasets. For the multi-modal dehazing method, training utilizes paired optical and SAR images from the datasets. We employ the same training and testing sets to evaluate both the comparison methods and the proposed method.

\begin{table*}[width=\textwidth]
\centering
\renewcommand{\arraystretch}{1.2}
\begin{center}
\caption{Quantitative comparison on MRSHaze and SEN12MS-CR datasets, reporting PSNR and SSIM metrics. \textcolor{red}{Red} indicates the best result among all methods, while \textcolor{blue}{blue} indicates the best result among the comparison methods.}
\begin{tabular}{c c c c c c c c c c}
\hline
\multirow{2}{*}{Model} & \multirow{2}{*}{Venue} & \multirow{2}{*}{G/S} & \multicolumn{2}{c}{MRSHaze} & & \multicolumn{2}{c}{SEN12MS-CR} & \multirow{2}{*}{Params (M)} & \multirow{2}{*}{FLOPs} \\
\cline{4-5} \cline{7-8}
 &  &  & PSNR$\uparrow$ & SSIM$\uparrow$ & & PSNR$\uparrow$ & SSIM$\uparrow$ & & \\
\hline
SwinIR~\cite{9607618}& ICCV'2021 & Single & 28.53 & 0.9025 & & 16.18 & 0.5103 & 7.77 & 2.03T \\
FFA-Net~\cite{qin2020ffa} & AAAI'2020 & Single & 30.33 & 0.9226 & & 17.57 & 0.5354 & 4.46 & 1.15T \\
MixDehazeNet~\cite{Mixdehazenet} & arXiv'2023 & Single & 30.70 & 0.9271 & & 17.78 & 0.5409 & 3.17 & 114.30G \\
TaylorFormer~\cite{10378631} & ICCV'2023 & Single & 28.95 & 0.9068 & & 15.59 & 0.4478 & 2.68 & 127.8G \\
C2PNet~\cite{Zheng_2023_CVPR} & CVPR'2023 & Single & 30.67 & 0.9263 & & 17.37 & 0.5309 & 7.17 & 1.84T \\
DEA~\cite{10411857} & TIP'2024 & Single & 30.38 & 0.9252 & & \textcolor{blue}{18.17} & 0.5628 & 7.79 & 78.19G \\
SF-Net~\cite{cui2023selective} & ICLR'2023 & Single & 32.54 & 0.9514 & & 17.89 & \textcolor{blue}{0.5816} & 13.27 & 497.76G \\
FS-Net~\cite{10310164} & TPAMI'2023 & Single & \textcolor{blue}{32.66} & \textcolor{blue}{0.9515} & & 17.87 & 0.5805 & 13.28 & 433.36G \\
FocalNet~\cite{Cui_2023_ICCV} & ICCV'2023 & Single & 31.69 & 0.9451 & & 17.83 & 0.5781 & 3.74 & 122.14G \\
DehazeFormer-s~\cite{10076399} & TIP'2023 & Single & 30.68 & 0.9271 & & 17.68 & 0.5325 & 1.29 & 47.95G \\
DehazeFormer-m~\cite{10076399} & TIP'2023 & Single & 31.15 & 0.9334 & & 17.83 & 0.5443 & 4.64 & 187.66G \\
VIF~\cite{YU2024128105} & arXiv'2024 & Guided & 30.89 & 0.9262 & & 16.99 & 0.5278 & 9.78 & 708.36G \\
UGSR~\cite{gupta2021toward} & TIP'2021 & Guided & 26.13 & 0.8561 & & - & - & 2.72 & 711.71G \\
MoG-DCN~\cite{dong2021model} & TIP'2021 & Guided & 29.57 & 0.9168 & & 16.68 & 0.5099 & 6.70 & 4.16T \\
GDEA~\cite{10411857} & TIP'2024 & Guided & 31.29 & 0.9338 & & 17.58 & 0.5573 & 21.30 & 411.14G \\
GFocalNet~\cite{Cui_2023_ICCV} & ICCV'2023 & Guided & 31.85 & 0.9452 & & 17.25 & 0.5777 & 23.17 & 465.60G \\
\midrule
\textbf{DehazeMamba-T} & \textbf{Ours} & \textbf{Guided} & \textbf{32.69} & \textbf{0.9507} & & \textbf{18.67} & \textbf{0.6151} & \textbf{6.65} & \textbf{180.72G} \\
\textbf{DehazeMamba-B} & \textbf{Ours} & \textbf{Guided} & \textbf{32.96} & \textbf{0.9540} & & \textbf{18.71} & \textbf{0.6101} & \textbf{10.37} & \textbf{341.24G} \\
\textbf{DehazeMamba-L} & \textbf{Ours} & \textbf{Guided} & \textbf{\textcolor{red}{33.39}} & \textbf{\textcolor{red}{0.9563}} & & \textbf{\textcolor{red}{19.04}} & \textbf{\textcolor{red}{0.6236}} & \textbf{17.87} & \textbf{634.77G} 
\\ \hline
\end{tabular}
\label{exp:quant}
\end{center}
\vspace{-0.3cm}
\end{table*}

\subsection{Quantitative and Qualitative Comparison}

\noindent\textit{1) Qualitative Results}: Fig.~\ref{visualization} illustrates the qualitative differences in restoration performance between DehazeMamba and comparative methods on the MRSHaze dataset. DehazeMamba achieves complete restoration of targets in heavily haze-affected regions. In the first example, other methods fail to fully recover obscured targets: FFA-Net and VIF partially restore outlines but miss critical details, while SwinIR, TaylorFormer, DEA, and DehazeFormer lose significant target information entirely. DehazeMamba, in contrast, accurately restores both object boundaries and fine-grained details. This superior performance pattern is consistent across multiple test samples. DehazeMamba's excellence stems from its effective utilization of SAR information to guide RGB image dehazing. In densely hazy regions, it strategically combines SAR-derived structural information with optical data to facilitate comprehensive target recovery. While VIF demonstrates some capability in restoring contours, DehazeMamba's sophisticated SAR-optical fusion architecture enables optimal exploitation of complementary information for superior dehazing results.

Fig.~\ref{cr1} presents qualitative comparisons on the SEN12MS-CR dataset. Single-image dehazing methods like SwinIR not only fail to effectively remove haze but also introduce significant distortions. Although the multi-modal VIF method occasionally recovers some texture information (as evident in the third row), its dehazed images still exhibit severe distortions. In contrast, DehazeMamba effectively addresses distortion issues and intelligently leverages SAR information to recover optical image textures, resulting in superior dehazing performance.

Fig.~\ref{real} showcases visual comparisons on real-world hazy images. All single-image methods exhibit substantial distortion artifacts. In regions with prominent SAR textures, multi-modal methods like MoG-DCN and VIF can generate clearer textures in corresponding optical image areas. However, these approaches suffer from two major limitations: first, the dehazed images often display poorly harmonized colors; second, in areas where SAR textures are less pronounced, introduced SAR information disrupts the original pixel composition, resulting in significant information degradation. DehazeMamba produces images with more natural color reproduction and selectively incorporates SAR information, effectively recovering haze-obscured targets while preserving information integrity in non-hazy regions.

\noindent\textit{2) Quantitative Results}: Table~\ref{exp:quant} presents a comprehensive quantitative performance comparison. DehazeMamba consistently outperforms all previous state-of-the-art methods on both MRSHaze and SEN12MS-CR datasets in terms of PSNR and SSIM metrics. Notably, DehazeMamba-L achieves a PSNR exceeding 33 dB on MRSHaze, surpassing the best comparative method, FS-Net, by a significant margin of 0.73 dB.

Existing multimodal methods (VIF, UGSR, MoG-DCN) demonstrate limited capability in handling complex remote sensing haze patterns, and their adoption of global fusion strategies introduces artifacts in haze-free regions, resulting in performance inferior to that of unimodal methods. On the MRSHaze dataset, the adapted multimodal approaches (GDEA, GFocalNet) show improvements but still fall short of DehazeMamba's performance. On the SEN12MS-CR dataset, adapted multimodal methods (GDEA, GFocalNet) exhibit poorer performance than their unimodal counterparts (DEA, FocalNet). This degradation is primarily attributable to the lack of denoising preprocessing in SEN12MS-CR's SAR images, where the introduction of SAR data containing substantial noise adversely affects optical image quality.

These results highlight the limitations of global guidance approaches and underscore the importance of selective SAR information integration. DehazeMamba's superior performance demonstrates its effectiveness in adaptively leveraging SAR information for guided optical image dehazing, particularly in addressing the challenges posed by non-uniform haze distribution patterns.

\begin{figure}[!t]
  \centering
   \includegraphics[width=\linewidth]{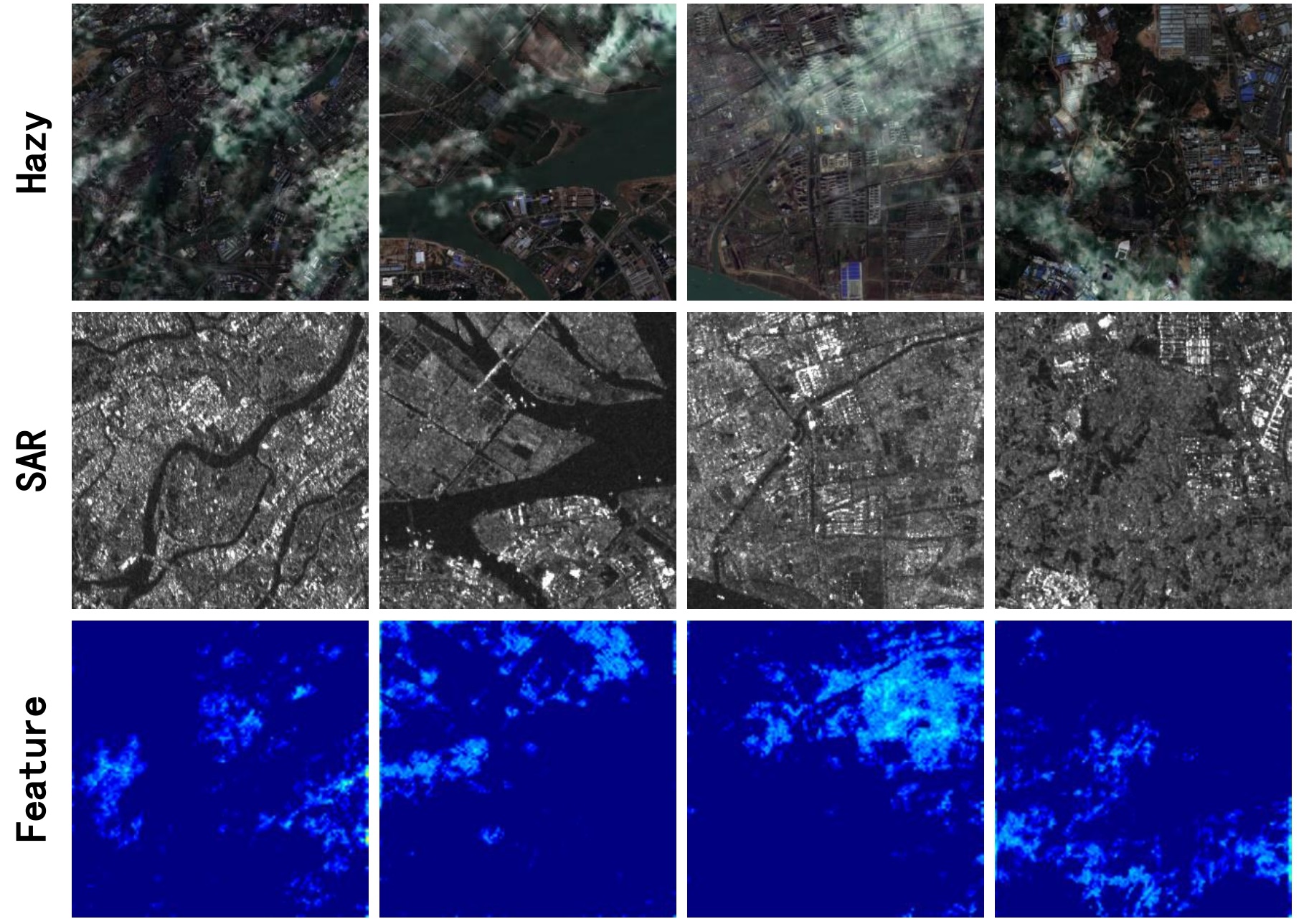}
   \caption{Visualization of HPDM's first-layer weight map ($W1$). Higher weights (brighter regions) directly correspond to haze-affected areas, with weight intensity proportional to haze density. This demonstrates HPDM's capacity to precisely detect both the spatial distribution and concentration of haze regions.}
   \label{hppm}
\end{figure}

\subsection{Visualization of HPDM}
To validate the efficacy of the Haze Perception and Decoupling Module, we visualize the weight map generated by the sigmoid activation function in its first layer. This map illustrates HPDM's capability to dynamically adjust SAR information integration based on haze distribution in optical images. As shown in Fig.~\ref{hppm}, there is a strong correlation between haze distribution in optical images and high-weight regions in the generated weight map. Denser haze areas correspond to higher weights, confirming HPDM's ability to accurately identify haze-affected regions and adaptively modulate SAR feature integration based on both haze density and spatial location. These visualizations substantiate HPDM's effectiveness in selectively incorporating SAR information, enhancing the dehazing process in haze-affected areas while preserving original details in clear regions. This adaptive mechanism significantly contributes to DehazeMamba's superior performance in handling non-uniform haze distributions characteristic of remote sensing imagery.

\subsection{Misguidance Phenomenon of Global Fusion}
\begin{figure*}
\centering
\includegraphics[width=\textwidth]{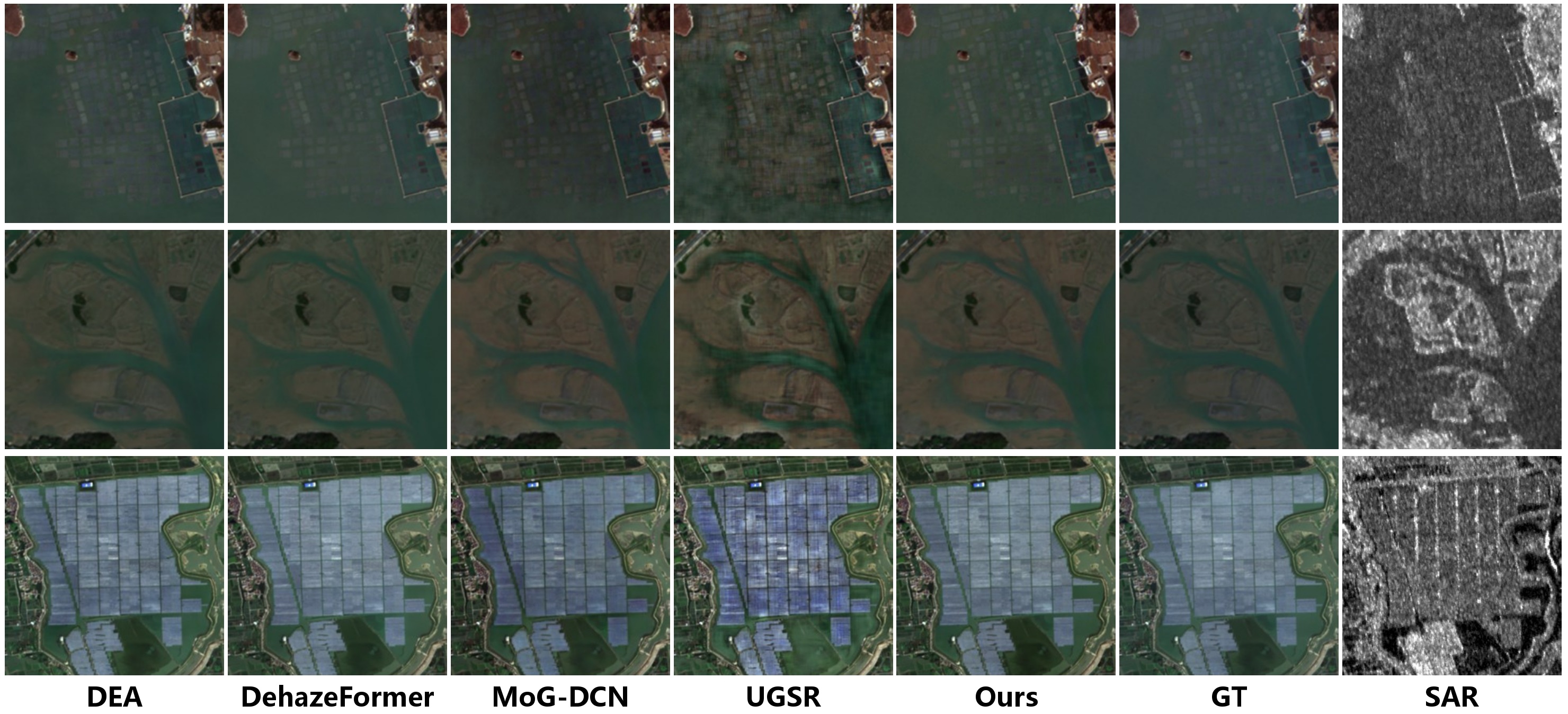}
\caption{Illustration of the misguidance phenomenon resulting from global fusion strategies. GT represents ground truth inputs. DEA and DehazeFormer employ single-modal approaches. MoG-DCN and UGSR implement SAR-optical fusion with global fusion strategies. Our DehazeMamba utilizes a selective local fusion approach that preserves image integrity while effectively integrating complementary information.}
\label{wrong_guide}
\end{figure*}

We identify a critical issue wherein global fusion strategies often introduce unnecessary SAR information into optical images, disrupting the original pixel composition in haze-free regions and consequently degrading overall image quality. Fig.~\ref{wrong_guide} illustrates the performance of three distinct dehazing approaches when processing negative samples (haze-free images). Single-modal methods (DEA and DehazeFormer) exhibit minimal alteration of the input images, reflecting the absence of interference from extraneous SAR information. In contrast, multimodal methods implementing global fusion strategies (MoG-DCN and UGSR) introduce substantial SAR information into optical images indiscriminately, resulting in color inconsistencies and visible artifacts. DehazeMamba, employing a selective local fusion strategy, effectively regulates the influence of SAR information on optical imagery, ensuring preserved quality in the processed results.

\begin{figure}[!t]
  \centering
   \includegraphics[width=\linewidth]{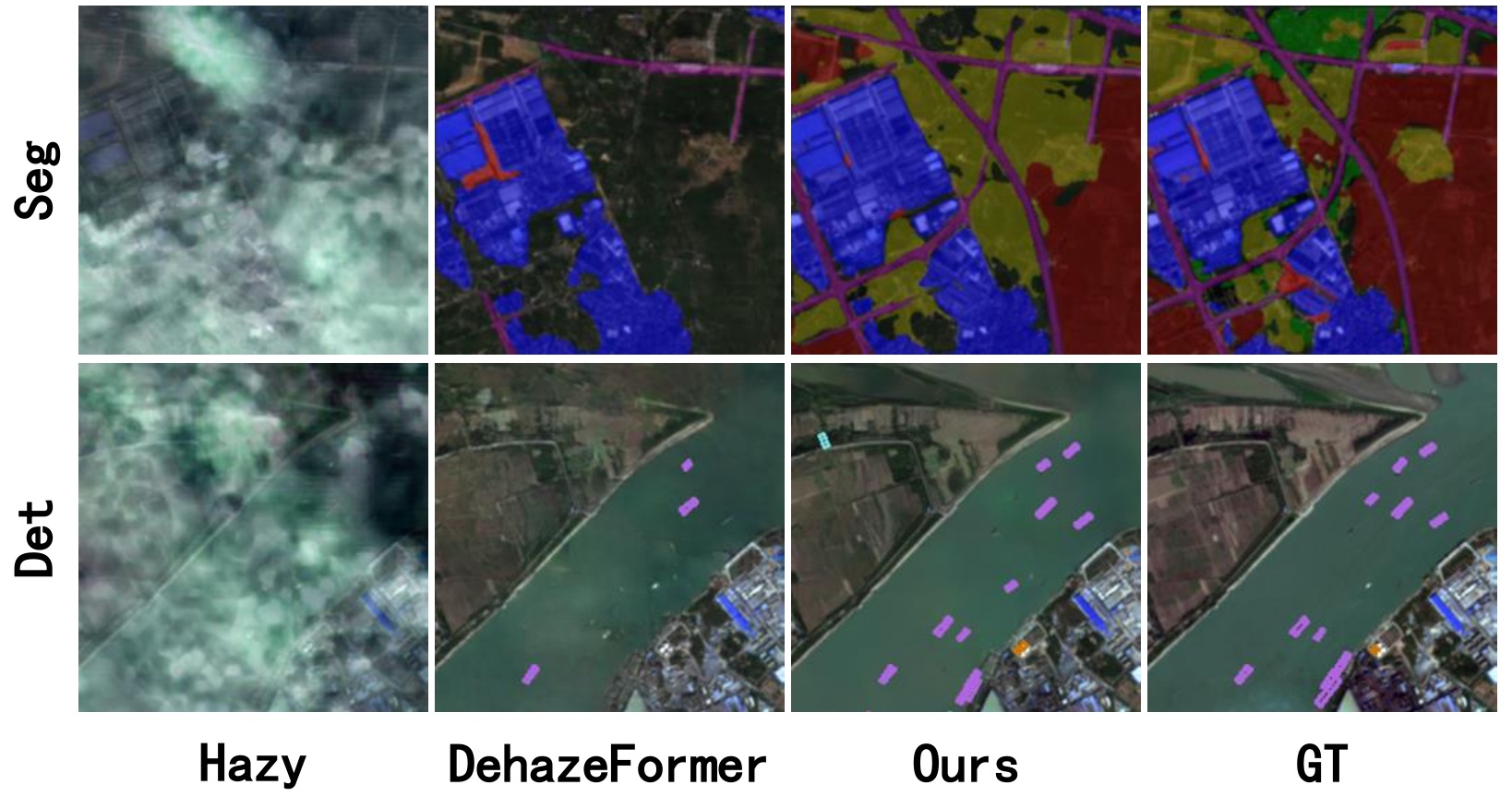}
   \caption{Comparison of semantic segmentation and object detection results after applying different dehazing methods. DehazeMamba produces dehazed images that enable more accurate downstream task performance, with results closely approximating those achieved using ground truth (GT) images. }
   \label{task}
\end{figure}

\subsection{Comparison on Downstream Tasks}
Fig.~\ref{task} demonstrates semantic segmentation and object detection outcomes obtained after processing dehazed images using ALE-SEG~\cite{xu2023analytical} and YOLOv8-x~\cite{varghese2024yolov8}, respectively. Images dehazed using DehazeFormer exhibit significant missing and false detections in both tasks. In contrast, DehazeMamba effectively integrates complementary optical and SAR information, substantially enhancing dehazed image quality and consequently producing segmentation and detection results that closely align with ground truth annotations. These findings underscore DehazeMamba's superior efficacy in supporting critical downstream applications such as semantic segmentation and object detection, confirming its practical utility in real-world remote sensing analysis workflows.

\begin{table}[!b]
\renewcommand{\arraystretch}{1.2}
\begin{center}
\caption{Ablation studies on HPDM (H) and PFM (P). When PFM is absent, pixel-wise addition is used for feature fusion.  }
\begin{tabular}{c c c c c c}
\hline
H & U & PSNR$\uparrow$ & SSIM$\uparrow$ & Params(M)  & GFLOPs\\
\hline
- & -  & 31.81 & 0.9438 & 4.62 & 147.42  \\
$\surd$ & -   & 32.02 & 0.9446  & 4.81 & 150.58 \\
 -  &  $\surd$  & 32.58 & 0.9501    & 6.46 & 177.56  \\
$\surd$ & $\surd$   & \textbf{\textcolor{red}{32.69}} & \textbf{\textcolor{red}{0.9507}}   & 6.65 &  180.72 \\ 
\hline
\label{ab:module}
\vspace{-0.5cm}
\end{tabular}
\end{center}
\end{table}

\begin{table}[!b]
\renewcommand{\arraystretch}{1.2}
\begin{center}
\caption{Ablation studies on the first (F) and second (S) phases of PFM.  When F is absent,  pixel-wise addition is used to generate coarse fused feature. }
\begin{tabular}{c c c c c c}
\hline
F  & S & PSNR$\uparrow$ & SSIM$\uparrow$ & Params(M)  & GFLOPs\\
\hline
- & -  & 32.02 & 0.9446 &  4.81 & 150.58  \\
$\surd$ & -   & 32.29 & 0.9468  & 5.14 & 156.01 \\
 -  &  $\surd$  & 32.51 & 0.9493   & 6.32 & 175.28  \\
$\surd$ & $\surd$   & \textbf{\textcolor{red}{32.69}} & \textbf{\textcolor{red}{0.9507}}   & 6.65 &  180.72   \\ 
\hline
\label{ab:stage}
\vspace{-0.5cm}
\end{tabular}
\end{center}
\end{table}

\subsection{Ablation Study}
Using the proposed MRSHaze dataset, we conduct comprehensive ablation studies on DehazeMamba-T to evaluate the contributions of key architectural components. As shown in Table~\ref{ab:module}, the absence of either HPDM or PFM results in significant performance degradation (PSNR: -0.11 dB and -0.67 dB; SSIM: -0.0006 and -0.0061, respectively), demonstrating the effectiveness of our strategy for selectively fusing optical and SAR information based on modality-specific uncertainty in haze-affected regions. Table~\ref{ab:stage} reveals that utilizing only the first or second stage of PFM also leads to measurable performance decline (PSNR: -0.40 dB and -0.18 dB; SSIM: -0.0039 and -0.0014), confirming that our two-stage progressive fusion approach more comprehensively and effectively integrates optical and SAR information.

We further validate the importance of SAR information in Table~\ref{ab:SAR}, which demonstrates that using only optical images results in substantial degradation (-1.41 dB PSNR and -0.0114 SSIM). This confirms that SAR information significantly contributes to optical image dehazing when appropriately integrated. Table~\ref{ab:loss} verifies the effectiveness of our dual-domain loss function, showing that employing only spatial or frequency domain loss significantly impairs dehazing performance (PSNR: -0.97 dB and -0.33 dB; SSIM: -0.0126 and -0.0006). This highlights the importance of multi-domain constraints for generating high-quality dehazed images, as single-domain supervision proves insufficient for effective network training.

\begin{table}[!t]
\renewcommand{\arraystretch}{1.2}
\begin{center}
\caption{Ablation study on SAR branch of DehazeMamba-T.  }
\begin{tabular}{c c c c c}
\toprule
 SAR & PSNR & SSIM & Params (M) & FLOPs \\
\toprule
 - & 31.28 & 0.9393 & 2.42 & 82.90G \\
 $\surd$ &  \textbf{\textcolor{red}{32.69}}  & \textbf{\textcolor{red}{0.9507}}   & 6.65 & 180.72G \\
\toprule
\label{ab:SAR}
\vspace{-0.5cm}
\end{tabular}
\end{center}
\end{table}

\begin{table}[!t]
\renewcommand{\arraystretch}{1.2}
\begin{center}
\caption{Ablation Study on Loss Function Components. $\mathcal{L}_{f}$ and $\mathcal{L}_{s}$ represent frequency and spatial domain losses, respectively. }
\begin{tabular}{c c c c c c}
\toprule
 $\mathcal{L}_{f}$ & $\mathcal{L}_{s}$ & PSNR & SSIM  \\
\toprule
 - & $\surd$ & 31.72 & 0.9381  \\
 $\surd$ &  - &32.36 & 0.9501  \\
 $\surd$ &  $\surd$ &  \textbf{\textcolor{red}{32.69}}  & \textbf{\textcolor{red}{0.9507}}    \\
\toprule
\label{ab:loss}
\vspace{-0.5cm}
\end{tabular}
\end{center}
\end{table}

\section{Conclusion}
In this work, we have presented DehazeMamba, a novel SAR-guided dehazing network that effectively leverages complementary SAR information to enhance detail recovery in haze-degraded optical remote sensing imagery. Our approach introduces two key innovations: a Haze Perception and Decoupling Module that dynamically identifies haze-affected regions, and a Progressive Fusion Module that mitigates domain shift through quality-aware feature integration. Additionally, we have developed MRSHaze, a comprehensive high-resolution benchmark dataset with precisely aligned SAR-optical image pairs. Extensive experimental evaluation demonstrates DehazeMamba's significant performance advantages over existing state-of-the-art methods, both in dehazing quality metrics and downstream application performance.

While DehazeMamba demonstrates superior capabilities in SAR-guided optical remote sensing image dehazing, future research directions include exploring more efficient SAR-optical fusion techniques to further reduce computational overhead, investigating the integration of additional complementary modalities beyond SAR, and extending the application scope to more diverse and challenging remote sensing scenarios. The MRSHaze dataset will continue to facilitate advancements in this important research domain.

\bibliographystyle{unsrt}

\bibliography{ref}

\end{sloppypar}
\end{document}